\documentclass[journal]{IEEEtran}
\usepackage{cite}
\usepackage{amsmath,amssymb,amsfonts}
\usepackage{algorithmic}
\usepackage{graphicx}
\usepackage{hyperref}       
\usepackage{booktabs}        
\usepackage{floatrow}
\usepackage{microtype} 
\usepackage{rotating,booktabs,multirow}
\usepackage{subcaption}
\usepackage{wrapfig,lipsum}
\usepackage{diagbox}
\usepackage{caption}
\usepackage{algorithm}
\usepackage[algo2e]{algorithm2e}
\usepackage{algorithmic}
\usepackage{soul}
\usepackage{tabularx}
\usepackage{multirow}

\usepackage{rotating,booktabs,multirow}

\ifCLASSINFOpdf
\else
\fi

\begin{document}
\title{Non-contact Real time Eye Gaze Mapping System Based on Deep Convolutional Neural Network}

\author{\IEEEauthorblockN{1\textsuperscript{st} Hoyeon Ahn}\\
\IEEEauthorblockA{\textit{School of Electrical Engineering and Computer Science} \\
\textit{Gwangju Institute of Science and Technology}\\
Gwangju, Korea \\
ajhoyeon@gist.ac.kr}
}

\markboth{Journal of \LaTeX\ Class Files,~Vol.~14, No.~8, August~2015}%
{Shell \MakeLowercase{\textit{et al.}}: Bare Demo of IEEEtran.cls for IEEE Journals}

\maketitle

\begin{abstract}
Human-Computer Interaction(HCI) is a field that studies interactions between human users and computer systems. With the development of HCI, individuals or groups of people can use various digital technologies to achieve the optimal user experience. Human visual attention and visual intelligence are related to cognitive science, psychology, and marketing informatics, and are used in various applications of HCI. Gaze recognition is closely related to the HCI field because it is meaningful in that it can enhance understanding of basic human behavior. We can obtain reliable visual attention by the Gaze Matching method that finds the area the user is staring at.
In the previous methods, the user wears a glasses-type device which in the form of glasses equipped with a gaze tracking function and performs gaze tracking within a limited monitor area. Also, the gaze estimation within a limited range is performed while the user's posture is fixed.
We overcome the physical limitations of the previous method in this paper and propose a non-contact gaze mapping system applicable in real-world environments. In addition, we introduce the GIST Gaze Mapping (GGM) dataset, a Gaze mapping dataset created to learn and evaluate gaze mapping.

\end{abstract}


\IEEEpeerreviewmaketitle

\section{Introduction}
\IEEEPARstart{T}{he} human eye is the most important source of information in the HCI field and can obtain brightness information, motion, and depth information. Humans rely on visual information for most of the information they get through their senses.
Understanding the human visual system with these characteristics is important in the field of visual HCI research \cite{0-11}.
The human eye has a unique movement pattern and physical characteristics. By analyzing these unique properties, we can grasp the concentration of human beings and the state of emotion. We conduct essential visual perception through the eyes and collect core information\cite{0-12}.
Research on gaze recognition has developed rapidly since the 1970s \cite{02-9}. The field of HCI gaze recognition has been mainly studied as a means to assist the disabled and has been developed as a specialized device for eye tracking \cite{02-13}. Gaze recognition has been subdivided into various fields such as gaze-based user interface and human cognitive status investigation \cite{02-17}\cite{02-18}\cite{02-19}, and research has been introduced to prevent driver's drowsiness by being applied to automobile Advanced Driving Assistance Systems (ADAS)\cite{02-00}.
Recently, with the continuous growth of wearable electronic devices and hardware performance, we have been able to analyze eye movements in detail, and the existing eye-tracking and recognition methods are performed by wearing glasses-type electronic devices \cite{wear}.  
Both eye areas are detected through the eye-piece camera installed in the device.
The final gaze can be traced to the location of the region of the pupil within both detected eye regions.
The pattern of the human eye and gaze implies an individual's needs, intentions, and state of mind as a process of understanding social interaction \cite{0-11}. However, most of the eye recognition studies so far have been performed while wearing glasses-type electronic devices \cite{02-00}. In other words, experiments were performed with the human being participating in the experiment aware of the gaze recognition test environment.
We must ensure that human beings participating in the experiment do not recognize themselves as experimental humans by wearing hardware devices.
   Through the non-contact method is possible to reliably extract the reaction variables of the experimental unit, and is also possible to create a natural experimental environment that eliminates the psychological factors of humans involved in the experiment. When performed in a non-contact method is possible to grasp the pattern of the natural gaze region in which the psychological factor which is human error is eliminated.

\begin{figure}[!t]
\centering
\includegraphics[width=\columnwidth, ,trim={0cm 0cm 0cm 0cm}]{{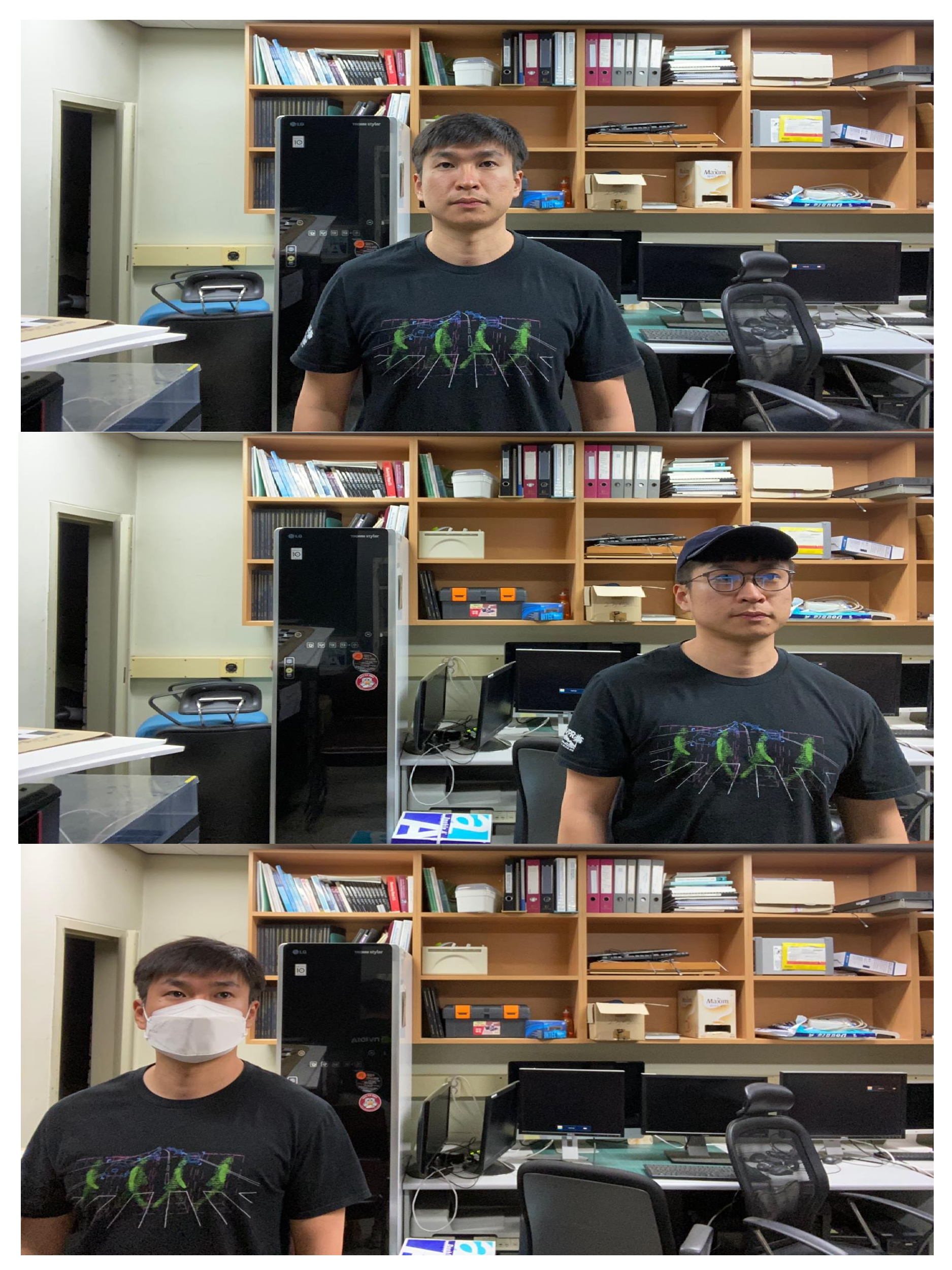}}
\caption {description of Shelf staring environment. The gaze environment varies depending on the user's glasses, masks, hats, etc., and the gaze habits and height of the user also vary.}
\label{fig:1}
\end{figure}

In order to achieve our goal, we must recognize the gaze without wearing the glasses-type device and be able to map the gaze regardless of the user's location and height.
However, the performance of the gaze recognition in an unconstrained environment without a wearable device is affected by external factors such as measurement distance, lighting, and whether the pupil region is captured. The coarse gaze area is acquired through head pose, and the fine final gaze area is mapped through the gaze recognition module.
The method proposed in this paper has the advantage of intuitively grasping the user's gaze without the glasses-type device in a non-contact manner and mapping the eye gaze regardless of the experimenter's location.

The proposed method consists of a three-step process. 1) Face module is performed (detection, alignment, and recognition). 2) The eye region is extracted from the face region and the gaze is estimated. 3) The final gaze is mapped to the target of the gaze panel by integrating the depth information obtained by the depth camera, gaze recognition information, and the head pose.
The main contributions of this paper are summarized as follows:
First of all, there is no public data for Gaze mapping, and a dataset GGM dataset was created to train the Gaze matching network. Also, the whole Gaze mapping system can be performed in a real environment using a Gaze estimator based on a deep convolutional neural network. We implemented the entire Gaze mapping system process by parallelizing and optimizing the system to run at the edge device NVIDIA TX2 with 7.5 fps performance.

The rest of the paper is organized as follows. Section 2 introduces related fields. Chapter 3 describes the Gaze Mapping System. Chapter 4 presents the experiment with our dataset and Section 5 describes the conclusion.

\section{Related work}
\subsection{Gaze Estimation}
Existing gaze recognition methods have been proposed mainly as hand-crafted features\cite{06-11}. After optimizing the fit of the hand-crafted feature model according to the linear regression equation, a method of estimating the final gaze method was studied \cite{06-25}. The feature-based gaze recognition can be implemented with a simple linear regression model. However, eye gaze feature is difficult to expect generalized performance in a real environment.
A model-based method has been proposed to show generalized performance improvement in a real environment \cite{06-30}\cite{06-39}. Modeling the eyeball technique calculates a gaze vector using feature points of the geometric model of the eye. Unlike feature-based gaze recognition which extracts local features of the eye region, models the entire eyeball area and recognizes the eye gaze. The eyeball modeling method uses high-dimensional input as a feature and learns the gaze mapping function.
Appearance-based gaze recognition is performed by image matching\cite{07-17}\cite{07-27} and modeling the entire eyeball. As for the method of modeling the eyeball, gaze recognition has been performed in the form of matching a model modeled in 3D with an eye image \cite{06-35}\cite{06-39}. The method based on 2D image matching is simple, but it has a disadvantage in that some cases show a sensitive response to changes in posture such as head pose and lighting, and affects the final gaze recognition performance.
On the other hand, information required for 3D shape modeling must be preceded by parameter measurements such as corneal radius and center, pupil radius, the distance between corneal center and pupil, angle, and refractive index between the optical axis and visual axis. The complex dataset is required, but the 3D model-based method shows more reliable gaze recognition results than an image matching method\cite{gazeml}.

Since 2012, deep neural networks with excellent performance in the computer vision field have been proposed \cite{alexnet}, and have been used in various computer vision tasks such as object classification and object detection \cite{09-16}\cite{inception}. In order to achieve a successful gaze recognition performance, learning to map from the eye image to the gaze direction must be performed well. Therefore, a large dataset for gaze recognition such as MPIIgaze\cite{mpiigaze} and RT-GENE \cite{RT-GENE} was proposed.
As the deep neural network is advanced, the deep neural network is also applied to the gaze recognition method. Some research shows superior gaze recognition performance compared to the gaze recognition method performed on feature-based methods \cite{06-38}\cite{06-14}.

\subsection{Face Detection, Alignment, and Recognition}
Face detection and recognition is a procedure that automatically finds a person's face in visual media and identifies them by individual IDs which is an essential and basic task in various face applications. The fundamental issues of computer vision such as occlusion, lighting changes, and pose changes, affect the performance of face recognition in the real environment.
Before the deep neural network was proposed, the face detection and recognition method based on the hand-crafted feature has been mainly studied. A cascade face detector using Haar-feature and Adaboost was proposed \cite{08-2}. Many studies have been proposed as a method capable of real-time processing and excellent performance in Occlusion and lighting change \cite{08-3}\cite{08-4}. As feature-based face research prospered, the main point of the core method developed into a deformable form that performs face recognition by modeling the relationship between face parts \cite{09-8}\cite{09-9}\cite{09-10}.

Deep neural network-based object detectors such as Yolo \cite{09-16} and SSD \cite{09-15} appeared, allowing face detection with excellent performance.
Both face detection and recognition are capable of end-to-end learning through a powerful deep learning optimization network, which has significantly changed the face research trends. Convolutional neural network (CNN) based method has the possibility of over-fitting as the network is deep and the number of parameters is large. Also, there is a disadvantage that it takes time to learn, but it has the advantage of facilitating generalization. Early CNN-based face recognition was used as an auxiliary task to improve the performance of face alignment \cite{08-22}. Since then, face alignment issues have been recognized by researchers as a major factor in detection and recognition performance, and research has been conducted to learn in a joint multi-task learning method \cite{mtcnn}.

As the performance of CNN is improved, many application developers possible to expect excellent detection and recognition performance for extremely small faces which not acquired in common surveillance environments such as WIDER FACE \cite{09-18}. A scale-invariant network was proposed to detect faces with different scales in each layer in a single network \cite{09-20}. A face recognition method using anchor-level attention, which has an excellent performance in occlusion cases, was proposed \cite{09-22}.

\begin{figure*}
\begin{center}
\includegraphics[width=\textwidth,trim={0cm 3cm 0cm 4cm}]{{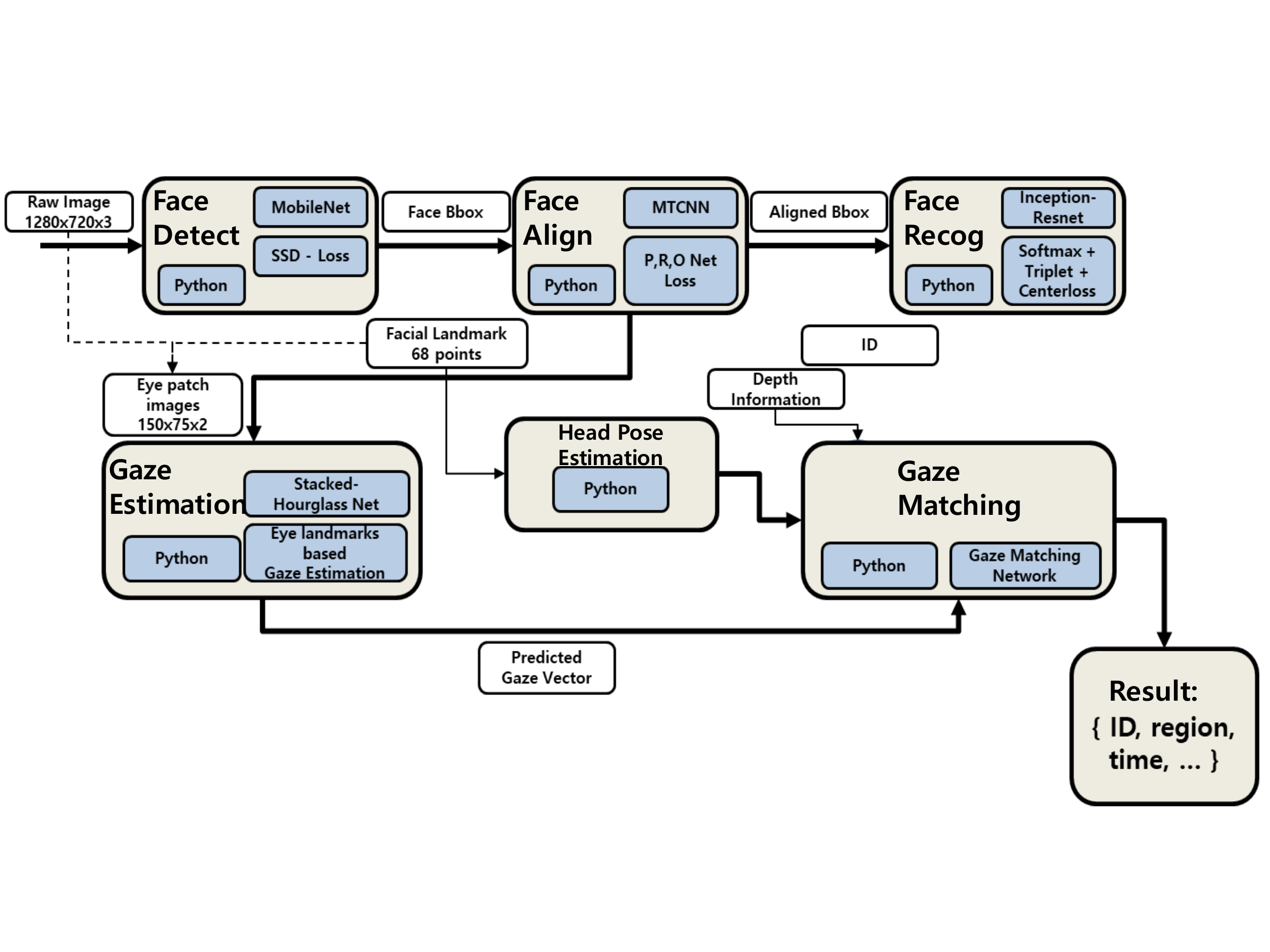}}
 \caption{Overall structure of Gaze Mapping System. The system consists of Face module, Gaze Estimation, and Gaze Matching module.
The Face module detects, aligns, and recognizes faces in order. The aligned face information is transferred to the Gaze Estimation module and finally used as the input of the Gaze Matching module along with the head pose and the predicted Gaze vector information.}
\label{fig:2}
\end{center}
\end{figure*}

\section{Eye Gaze Mapping System}
The proposed system recognizes users accessing the shelves. At the same time, pupil detection, gaze recognition, and head pose are performed within the recognized user's face area.
Finally, gaze mapping without a wearable device for gaze recognition can be performed by receiving gaze information, head pose, and depth information.
Since the performance of the Gaze Mapping module is dependent on the face detection and alignment module, each module was optimized.

In addition, In order to implement a real-time Gaze Mapping System that satisfies the resource limit of  NVIDIA Jetson TX2, we should carefully consider the network design and number of parameters of each module comprehensively.

The following subsections describe the detailed implementation of the face (detection, alignment, and recognition), gaze recognition, and gaze mapping modules. The last section describes how to implement  parallelized and optimized tasks assigned to each module.

\begin{figure}[!t]
\centering
\includegraphics[width=\columnwidth, trim={0cm 3cm 0cm 1cm}]{{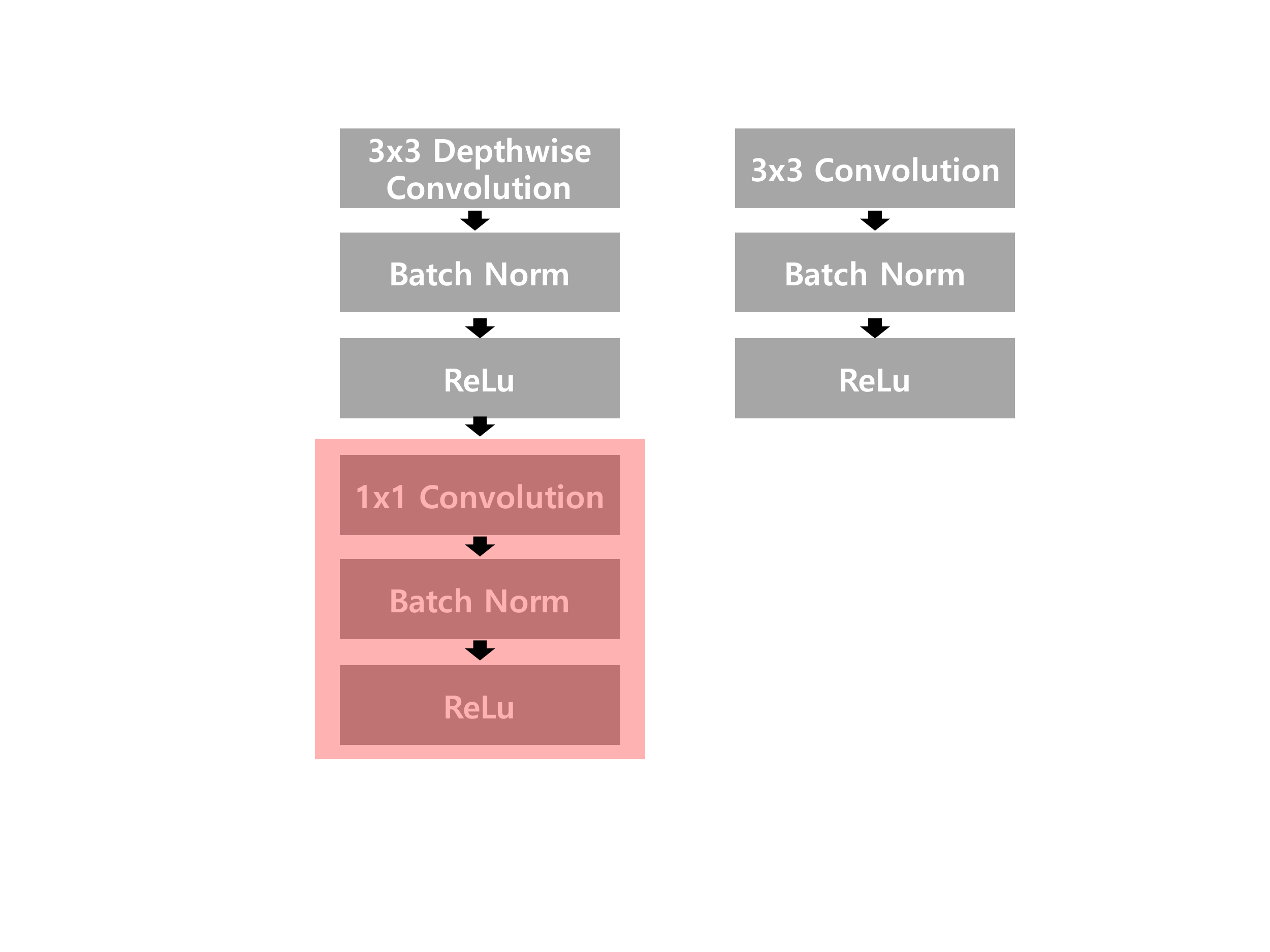}}
\caption {The core module of MobileNet is Depthwise Conv and then BN ReLU. Depthwise Conv reduces the number of parameters and computational cost than the normal convolution 2d process. Reft: Depth wise convolution structure. Right: Basic convolution layer structure.}
\label{fig:3}
\end{figure}

\subsection{Face Detection, Alignment, and Recognition}
We deployed the entire gaze mapping system to Edge devices. Therefore, face detection and alignment suitable for low power computing was adopted.
The face detection module suitable for the purpose of the real-time gaze mapping system is modified and applied to MobileNet \cite{mobilenet}, which is mainly used for edge device computing.

When ${D}_{K}$ is the kernel size, ${D}_{F}$ is the input channel size, M is the input channel, and N is the output channel, the general convolution computation is as shown in Eq (1) below.

\begin{equation}
{D}_{K}{\cdot}{D}_{K}{\cdot}{M}{\cdot}{N}{\cdot}{D}_{F}{\cdot}{D}_{F}
\label{equ:1}
\end{equation}

On the other hand, when the depthwise factorized Depthwise Separable Convolution of MobileNet is applied, the calculation amount is as follows.

\begin{equation}
{D}_{K}{\cdot}{D}_{K}{\cdot}{D}_{F}{\cdot}{D}_{F}{\cdot}{M}{+}{D}_{F}{\cdot}{D}_{F}{\cdot}{M}{\cdot}{N}
\label{equ:2}
\end{equation}
According to the eq (1) and (2) above, the ratio of the calculation amount is reduced as compared with the general convolution.

\begin{equation}
\frac{1}{N}+\frac{1}{{D}^{2}_{K}}
\label{equ:3}
\end{equation}

Assuming the kernel size is 3, the MobileNet-based face detector can achieve approximately 8 times the computational efficiency and effectively reduce the memory weight of the face detection module in the proposed gaze mapping system.

\begin{equation}
{D}_{K}{\cdot}{D}_{K}{\cdot}{\beta}{D}_{F}{\cdot}{\beta}{D}_{F}{\cdot}{\alpha}{M}{+}{\beta}{D}_{F}{\cdot}{\beta}{D}_{F}{\cdot}{\alpha}{M}{\cdot}{\alpha}{N}
\label{equ:4}
\end{equation}

In addition, as shown in Eq (4), when applying the final calculation, Width and Resolution multipliers $\alpha$ and $\beta$ are applied to reduce the face recognition module to be suitable for limited hardware NVIDIA jetson tx2 resources and deployable. A thin network can be created according to the value of $\alpha$. By adjusting $\beta$, the size of the input image and all internal layers can be reduced by the same ratio.
	The Face Alignment task is closely related to the detection task. Therefore, it is possible to maximize the effectiveness of sorting when performing learning with multitasking.

MTCNN \cite{mtcnn} proposed by zhang et al. consists of three models: P-Net, R-Net, and O-Net, and consists of a cascade inference structure. The network structure is designed to learn classification, landmark localization, and multi-task loss of box regressor in a joint learning manner. 

The main characteristic of MTCNN is implemented in the form of an image pyramid, which can be expected to improve detection and recognition performance by aligning faces of various sizes. In particular, there is leverage to maximize the performance of the gaze recognition module, which is dependent on face detection and alignment performance. 
The face recognition module applied Inception-ResNet\cite{inception-resnet}. In this paper, the effect of the residual module\cite{resnet} has the effect of reducing the convergence speed when learning a large data set. When performing face recognition, it should be recognized as different features, but Triplet loss \cite{facenet} for learning the distinguishing power of features with similar texture information is defined as in the following eq (5).

\begin{equation}
{L}={max}(0, \alpha + {d}({f}({X}_{i})), {f}({X}_{pos}))-{d}({f}({X}_{i}), {f}({X}_{neg})))
\label{equ:5}
\end{equation}

Here, $f$ is an embedding function and $d$ is a distance function to measure the distance between two inputs.
The embedding distance of the data ${X}_{pos}$ similar to the anchor as the reference point was learned to be more than a distance from ${X}_{neg}$ and the distance function was trained using $L2$ distance for the task.

\subsection{Gaze Estimation and Head Pose Estimation}
The hand-crafted features and model-based gaze recognition methods tend to be sensitive to lighting changes, resolution and occlusion. Therefore, these methods are difficult to apply in all real-world images.
Based on the stacked hourglass method, Park et al. \cite{gazeml} extracts eye landmarks by using features that capture necessary information across multiple scales.

Spatial information can be maintained by using only one skip layer for each scale. When we learn UnityEye \cite{unityeye} using these advantages, we can perform gaze recognition suitable for the real environment.
	Head pose estimation has been added to compensate for the instability of the Gaze estimation.
In order to estimate the head pose, some 2D coordinate information is required, and 3D coordinates of the corresponding 2D feature point are required.
In addition, since the 3D world is projected as a 2D image, a camera calibration process is required to remove the parameters inside the camera when calculating the 2D coordinates back into 3D coordinates. The head pose is estimated using 3D coordinates and camera matrix of OpenCV dlib 68 Facial Landmarks.

\subsection{Gaze Matching Network}
Gaze mapping System has a problem that is difficult to generalize because each person has different habits of gazing, such as side look and head angle. Also, few shot learning is suitable because the GGM dataset does not have many training images. Siamese-network, which takes two inputs and returns the similarity between two vectors, as shown fig. 4.

\begin{figure}[!t]
\centering
\includegraphics[width=\columnwidth, trim={0cm 0cm 0cm 0cm}]{{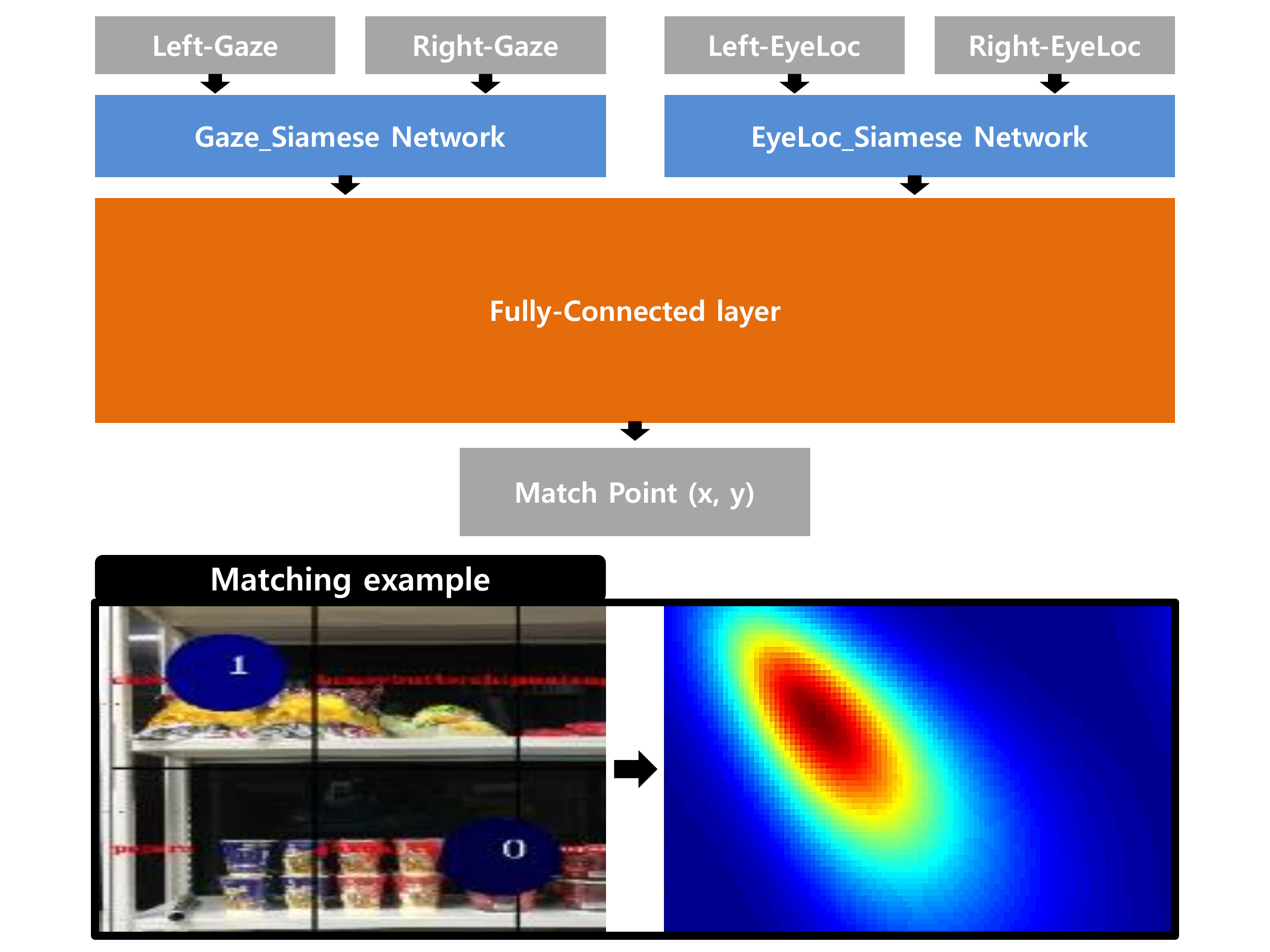}}
\caption {Description of Gaze Matching module. Gaze and eye position information of both eyes are used as input to each Siamese Network. The final matching point is calculated through the Fully-Connected layer.}
\label{fig:4}
\end{figure}

In order to identify feature extraction from left and right eye, siamese network is applied to location and gaze vector. Since the difference in scale of each input is very large, normalize each input value.
However, in case of depth, maximum and minimum values cannot be specified, so an additional 1-channel fully-connected layer is added. Gaze vector and Gaze location can be shifted to a similar scale.

\begin{figure}[!t]
\centering
\includegraphics[width=\columnwidth, trim={0cm 3cm 0cm 0cm}]{{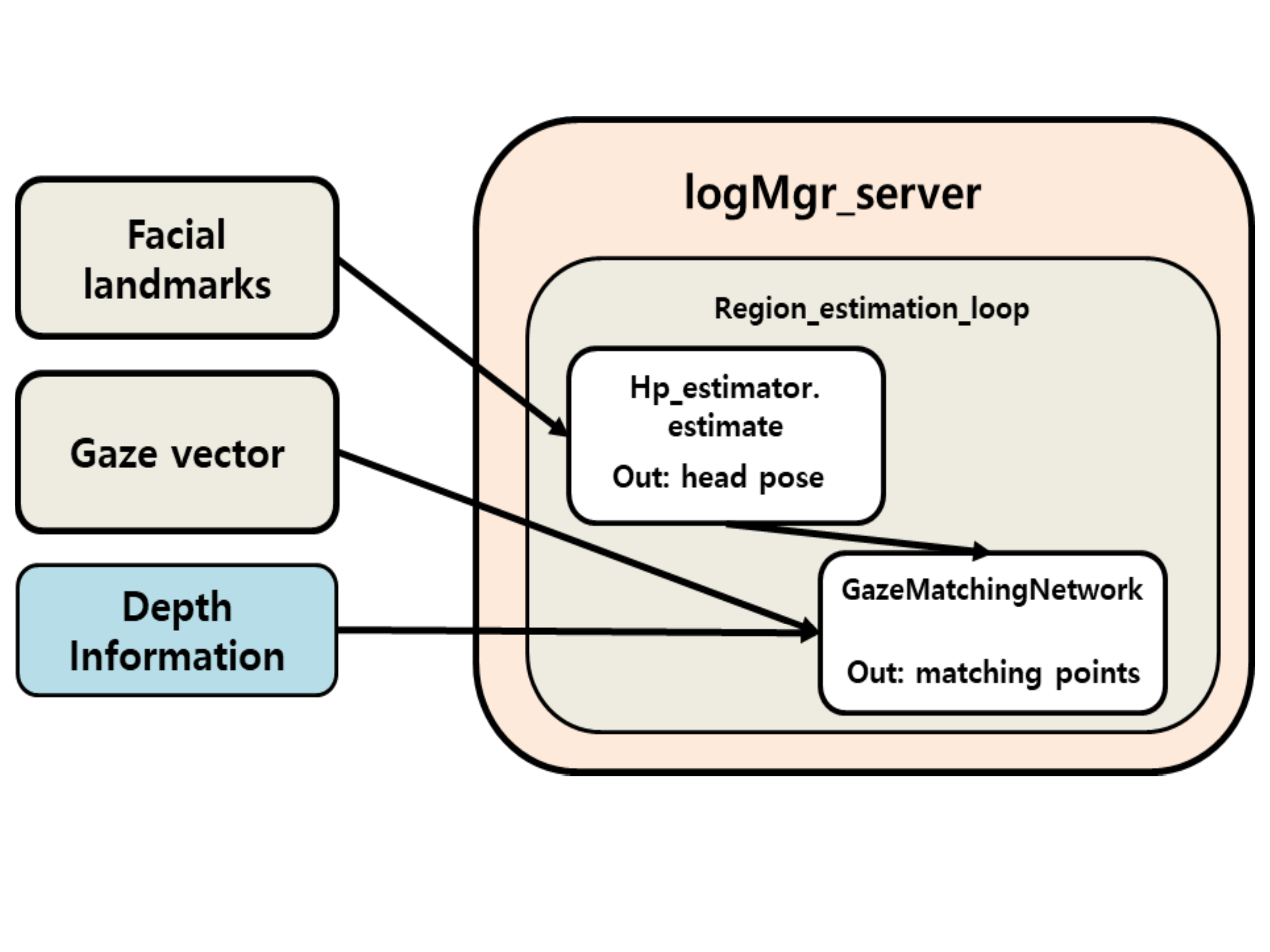}}
\caption {Gaze Matching module working structure. The module consists of a total of three sub-modules. Head pose estimator extracts the head pose vector using bounding boxes and landmarks collected from Jetson TX2. Gaze Matching Network extracts the gaze area within the shelf by combining the head pose vector, gaze vector, bounding box, and landmark. logMgr logs the events that occurred in each module and record them in the database.}
\label{fig:5}
\end{figure}

There are 3 sub-modules of Gaze mapping module (see fig.5).  
The head-pose estimator extracts the head-pose vector using information collected from Jetson tx2 (Bbox, Landmarks, etc.).
The Gaze-matching Network extracts the gaze area within the shelf by using the information (head pose vector, gaze vector, Bbox, landmark) collected in the module.
Logger and Visualization logs the events that occurred in each module and records them in the database and visualizes the logged data and transmitted images.

\begin{figure}[!t]
\centering
\includegraphics[width=\columnwidth, trim={0cm 2cm 0cm 0cm}]{{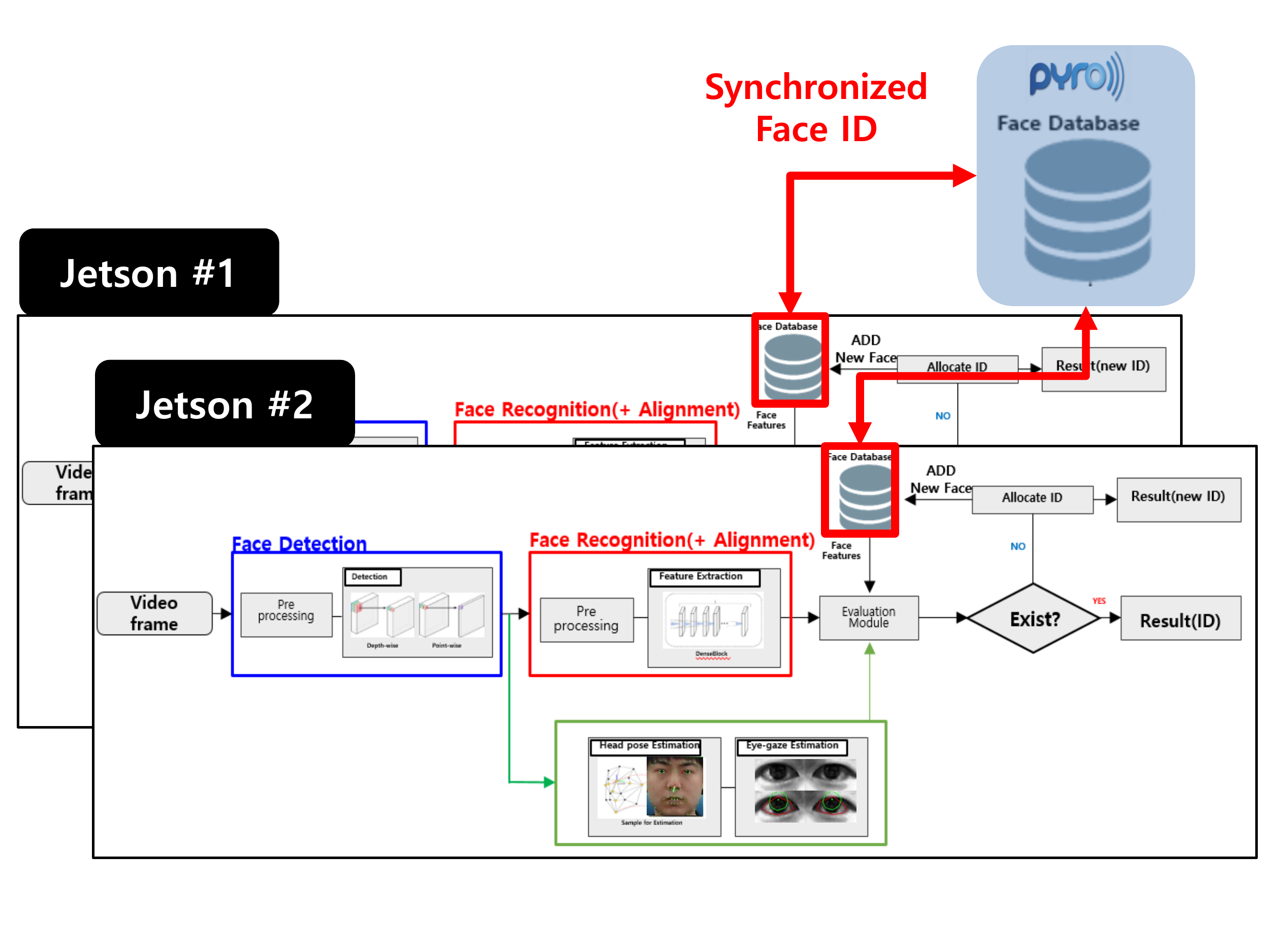}}
\caption {PyRO: A library that enables Python obect on a remote machine by utilizing the Python Remote Object. Face ID is shared by DB server in each Jetson TX2.}
\label{fig:7}
\end{figure}

\subsection{Gaze Mapping System parallelization and optimization}

The Gaze Mapping system consists of modules for face detection, alignment and recognition, and gaze estimation, and is implemented in parallel with queue and multi-process based to maximize resource utilization.
We implemented to allocate the CPU resource according to the task load of each module to adjust the overall inference speed.

\begin{figure}[!t]
\centering
\includegraphics[width=\columnwidth, trim={0cm 3cm 0cm 0cm}]{{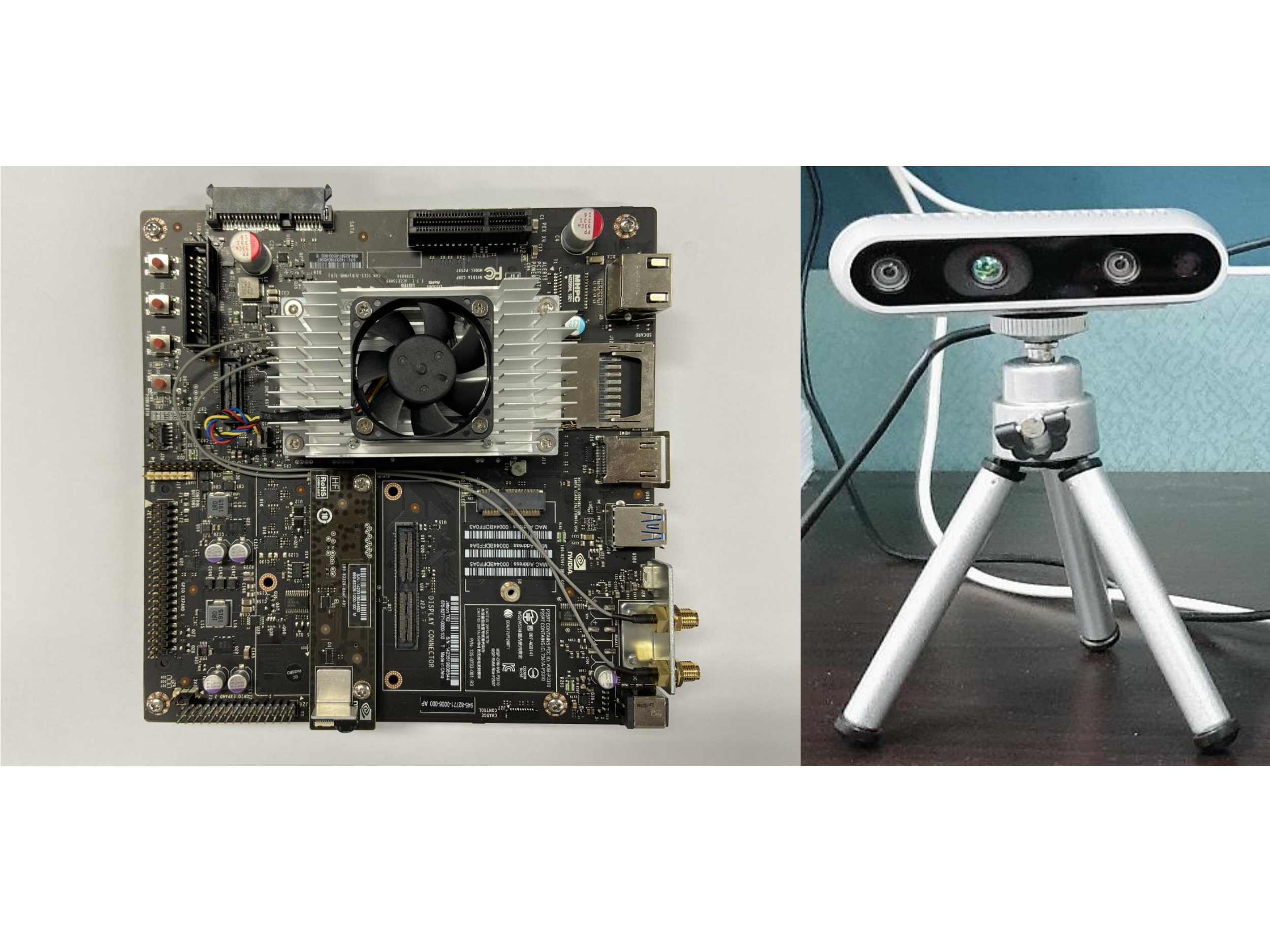}}
\caption {The Gaze Mapping System hardware configuration. The camera uses Intel Realsense D435, which can acquire depth information, and the board uses NVIDIA's Jetson TX2 board.}
\label{fig:7}
\end{figure}

Also, in order to share user IDs between different devices (see fig. 6), ID information was synchronized in real time using the Python Remote Object (PYRO).
All modules included in the Gaze mapping system are optimized using TensorRT, a platform for optimizing inference in deep learning models. In CNN, convolution, bias and ReLU layers can be combined into one CBR layer to increase memory efficiency and computation speed. 45 times faster in INT8 and FP16 precision than the CPU-only implementation.

\begin{figure}[!t]
\centering
\includegraphics[width=\columnwidth, trim={0cm 0cm 0cm 0cm}]{{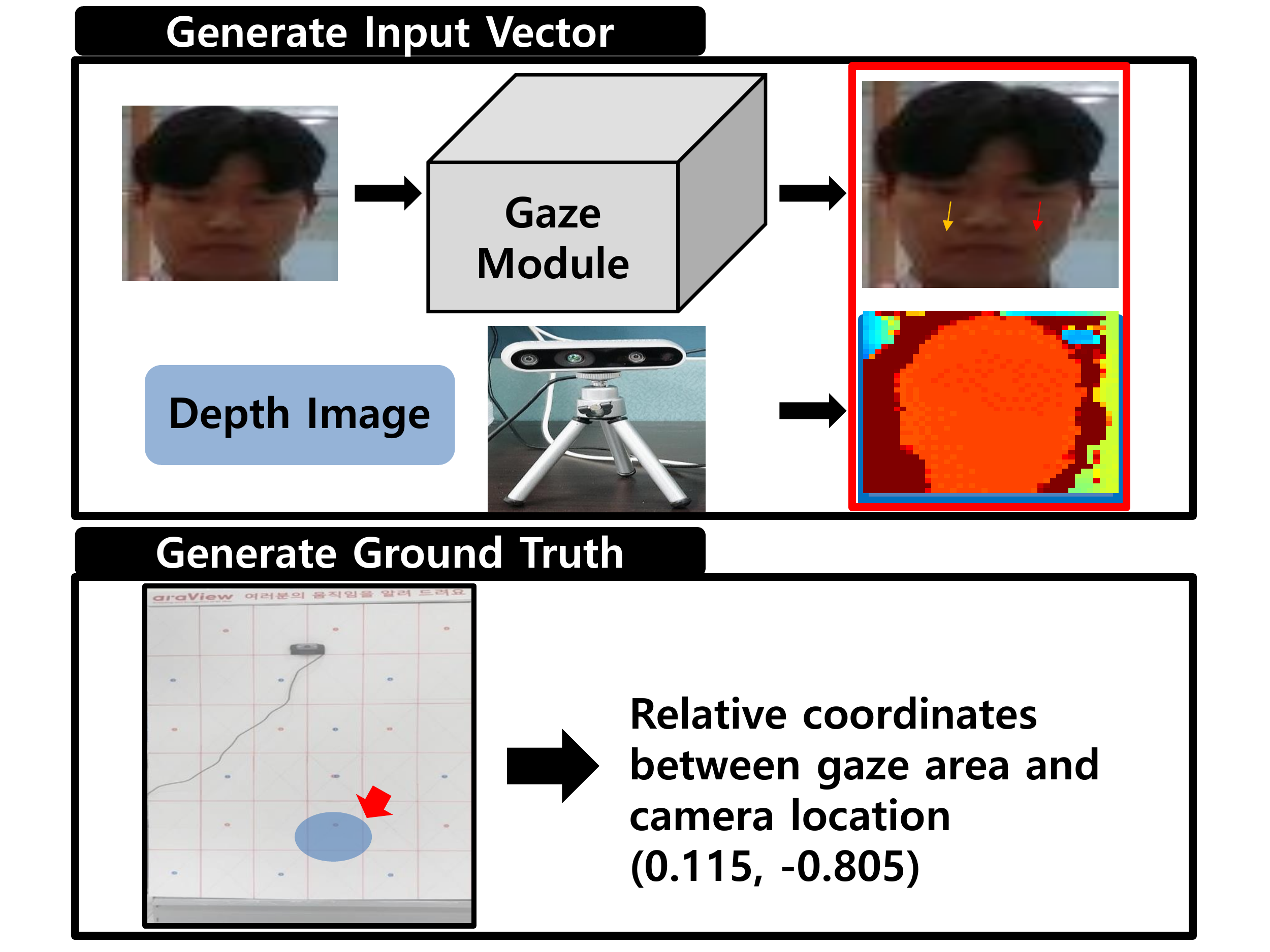}}
\caption { Description of creating a Gaze Matching network dataset. The gaze module output gaze vector and the image depth information provided from the depth camera are made into a single vector, and the final ground truth is produced from the user's gaze area and relative coordinates with the camera.}
\label{fig:7}
\end{figure}

\section{Experiments}
\subsection{GIST Gaze Mapping dataset}
GIST Gaze Mapping dataset was produced by generating the test environment in a total of 9 cases by changing the gaze area distance (0.75m, 1m, 1.5m) and the user's relative position from the camera at 0.5m intervals (left, right, center).

\begin{figure}[!t]
\centering
\includegraphics[width=\columnwidth, trim={0cm 0cm 0cm 0cm}]{{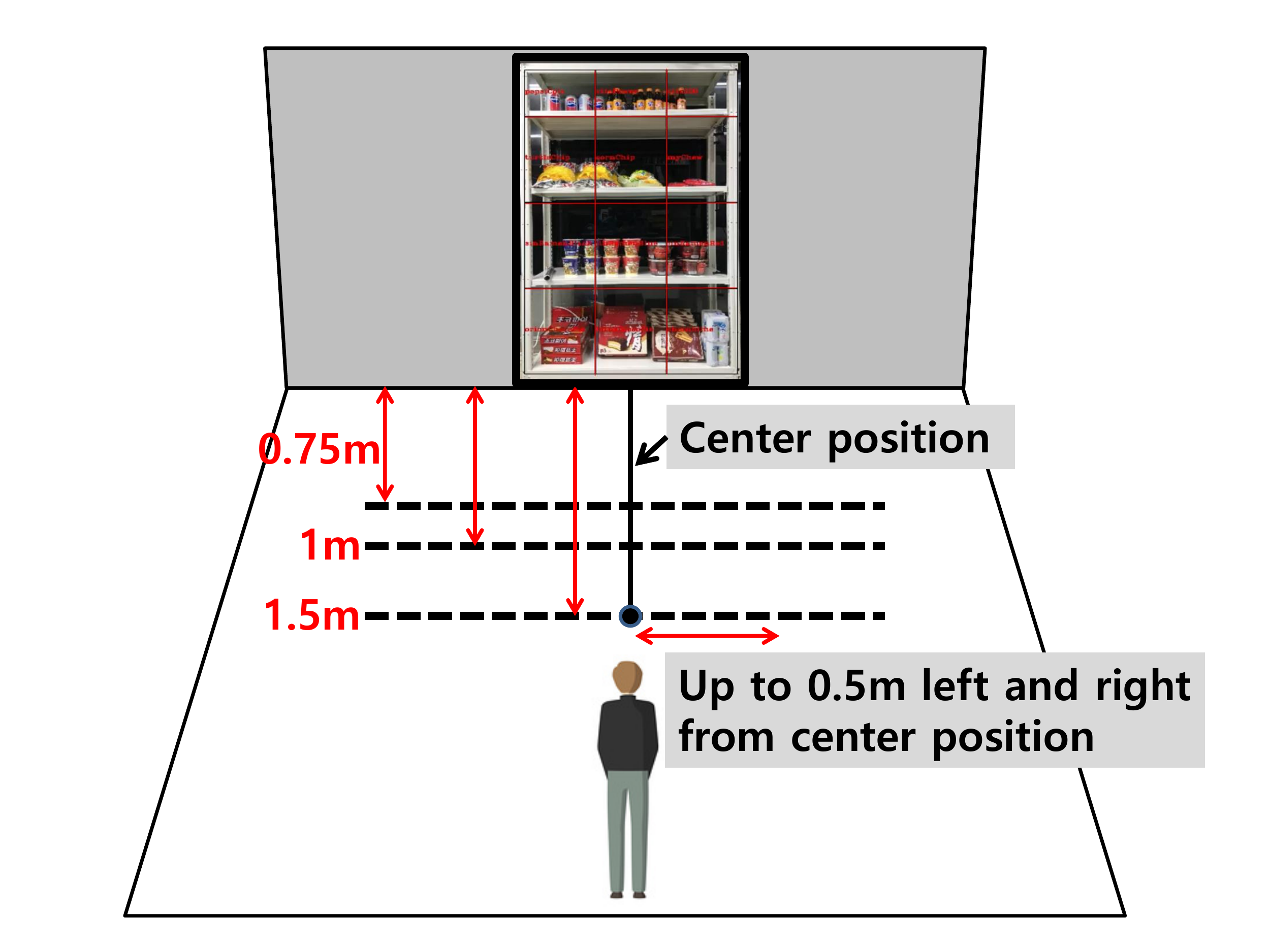}}
\caption {Description of GIST Gaze Mapping dataset acquisition method and experimental environment. The user gazes at a specific area (6x6 grid) in front of the shelf.}
\label{fig:7}
\end{figure}

  The user's appearance in the GGM dataset changes while wearing glasses, hats and masks. Includes a dataset that looks sideways up to $30\,^{\circ}$ left and right.
The shape of the gaze panel is composed of 6x6 grids as shown in Fig. 10, and is divided into 36 sections.
The size of each grid is 0.17m x 0.23m, and the size of the entire panel is 0.9m x 1.38m. The camera is installed at a height of 1.5 m from the ground.

\begin{figure}[!t]
\centering
\includegraphics[width=\columnwidth, trim={0cm 0cm 0cm 0cm}]{{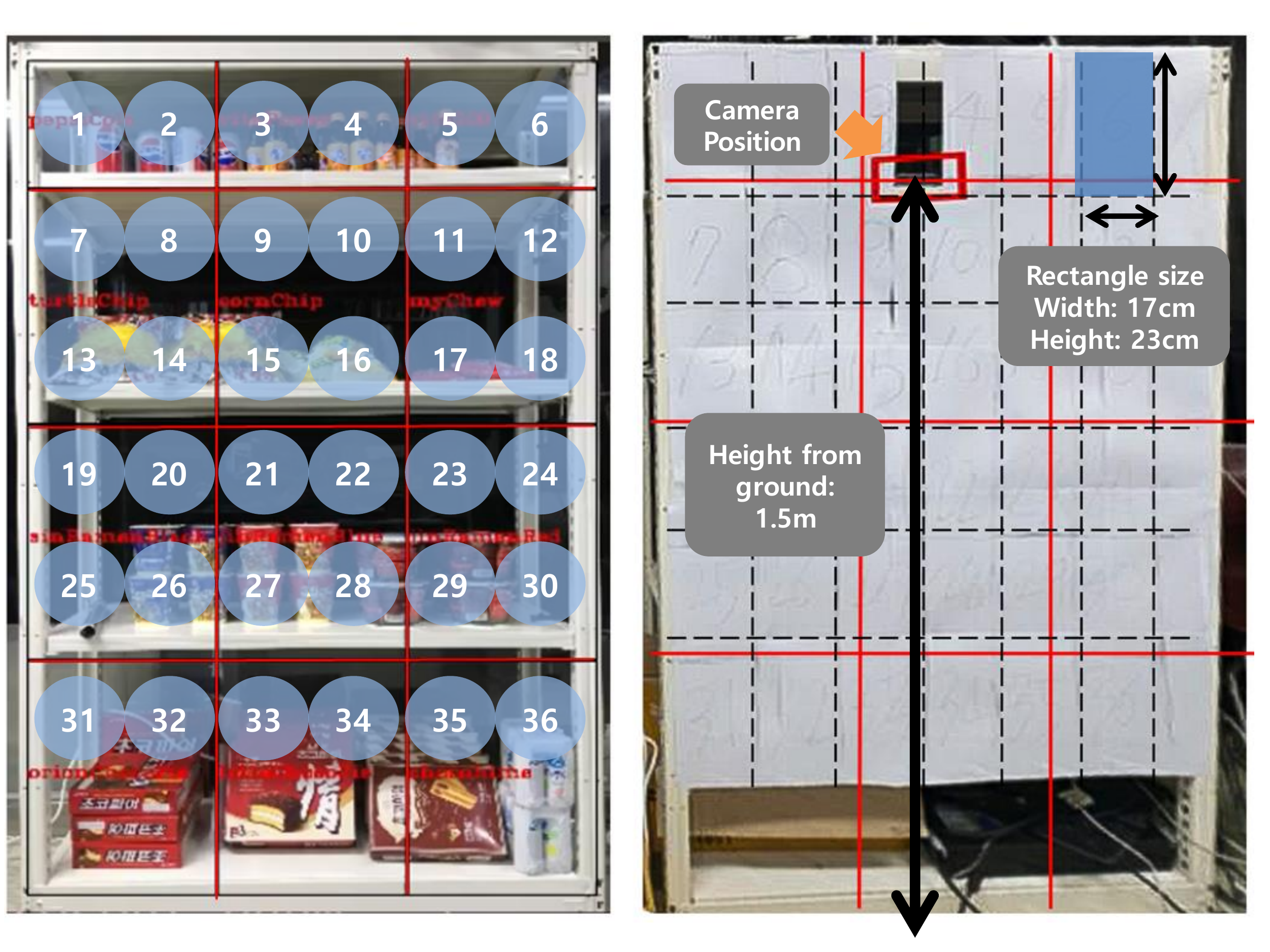}}
\caption {GIST Gaze Mapping System shelf configuration. The grid area is composed of a total of 36, and the shelves used in the actual distribution industry are used. Each grid is 17cm wide and 23cm wide, and the camera height is 1.5m from the ground.}
\label{fig:7}
\end{figure}

\subsection{Results}
A total of 90,000 GGM data training sets consisting of 5 users were trained for Gaze Mapping System. The evaluation of the Gaze mapping system was marked as correct when the user gazes for a specific area for 5 seconds, and the calculated value of the Intersection of Union (IoU) of the Gaze mapping prediction area and each grid area is 0.5 or more.

The test environment is divided into a total of 3 user cases as shown in Table 1, and experiments were conducted that included changes in user height and accessories.

When the user gazes at the lower area of the shelf (Grid number 25 to 36), we found that the performance of gaze mapping is lower than when gazing at the rest of the grid area.
This tendency is that when the user looks at the top of the shelf, region of interest is relatively easy to acquire from the whore eyeball area. Conversely, when looking down, Gaze estimation is limited due to insufficient pupil region acquisition.
When the user's gaze estimation on the shelf is incomplete, the result confirms that the approximate gaze can be guaranteed through the head pose information.
In the case of two users (USER CASE 1 and 2) according to the Test Definition in Table 2, evaluation was also performed, and processing is similar to the situation of one user.
The device used to implement the Gaze Mapping System is NVIDIA Jetson TX2. We tested at the same manufacturer's product, AGX Xavier, and showed a speed of 10 fps as shown in Table 3.

\begin{table*}
\caption{GIST Gaze Mapping dataset test scenario configuration. The user's case was constructed by changing the user's height, accessories, and face pose.
\label{with_encoder}}
{\begin{tabular*}{40pc}{@{\extracolsep{\fill}}llll@{}}
\toprule
CASES & Assumption  \\\hline
\midrule
USER CASE 1 & 1. The height of the two users must be different (User's height $155cm, 175cm, \pm 5cm$) \\  & 2. A/B/C/D Sub-CASE test(total number of points per user 80 = 10 points * 4 cases * 2 times) \\ & -A: No accessories, stare at the front \\ & -B: No accessories, $30\,^{\circ}$ side view \\ & -C: Wear accessories (glasses, hats, masks), stare at the front \\ & -D: Wear accessories (glasses, hats, masks), stare at $30\,^{\circ}$ from the left and right \\ &  3. Comply with Grid setting value (6x6 grid shelf) \\
\midrule
USER CASE 2 & 1. The height of the two users must be different ($User's height 165cm, 185cm, \pm 5cm$) \\  & 2. A/B/C/D Sub-CASE test(total number of points per user 80 = 10 points * 4 cases * 2 times) \\ & -A: No accessories, stare at the front \\ & -B: No accessories, $30\,^{\circ}$ side view \\ & -C: Wear accessories (glasses, hats, masks), stare at the front \\ & -D: Wear accessories (glasses, hats, masks), stare at $30\,^{\circ}$ from the left and right \\ &  3. Comply with Grid setting value (6x6 grid shelf) \\
\midrule
USER CASE 3 & 1. Turn the head as far as possible and stare \\  & 2. A/B/C/D Sub-CASE test(total number of points per user 40 = 10 points * 4 cases) \\ & -A: No accessories, stare at the front \\ & -B: No accessories, $30\,^{\circ}$ side view \\ & -C: Wear accessories (glasses, hats, masks), stare at the front \\ & -D: Wear accessories (glasses, hats, masks), stare at $30\,^{\circ}$ from the left and right \\ &  3. Comply with Grid setting value (6x6 grid shelf) \\
\bottomrule
\end{tabular*}}{}
\end{table*}

\begin{table*}
\caption{Experiments are performed according to the defined test environment. Various tests were performed depending on the number of users and the distance from the user's shelf.
\label{with_encoder}}
{\begin{tabular*}{40pc}{@{\extracolsep{\fill}}llll@{}}
\toprule
Test Definition & CASES & \# of trials & Accuracy(\%) \\\hline
\midrule
2 users stare at a specific point (0.75m)  & USER CASE 1 & 80 & 93.75 \\
\midrule
2 users stare at a specific point (0.75m)  & USER CASE 2 & 80 & 90.00 \\
\midrule
2 users stare at a specific point (1m)  & USER CASE 1 & 80 & 96.25 \\
\midrule
2 users stare at a specific point (1m)  & USER CASE 2 & 80 & 91.25 \\
\midrule
2 users stare at a specific point (1.5m)  & USER CASE 1 & 80 & 96.25 \\
\midrule
2 users stare at a specific point (1.5m)  & USER CASE 2 & 80 & 91.25 \\
\midrule
1 user stare at a specific point (0.75m)  & USER CASE 3 & 40 & 95.00 \\
\midrule
1 user stare at a specific point (1m)  & USER CASE 3 & 40 & 92.50 \\
\midrule
1 user stare at a specific point (1.5m)  & USER CASE 3 & 40 & 90.00 \\
\bottomrule
\end{tabular*}}{}
\end{table*}

\begin{table*}
\caption{A comparison of speculation and performance (fps) on Jetson TX2 and Jetson Xavier  \label{with_encoder}}
{\begin{tabular*}{40pc}{@{\extracolsep{\fill}}llll@{}}
\toprule
Hardware & Jetson AGX Xavier & Jetson TX2 \\\hline
\midrule
CPU(ARM)  & 8-core Carmel ARM CPU @ 2.26 GHz & 6-core Denver and A57 @ 2 GHz	\\
\midrule
GPU  & 512 Core Volta @ 1.37 GHz & 256 Core Pascal @ 1.3 GHz \\
\midrule
Memory  & 16 GB 256-bit LPDDR4x @ 2133 MHz & 8 GB 128-bit LPDDR4 \\
\bottomrule
speed  & 10 fps & 7.5 fps\\
\bottomrule
\end{tabular*}}{}
\end{table*}

\begin{figure}[!t]
\centering
\includegraphics[width=\columnwidth, trim={0cm 0cm 0cm 0cm}]{{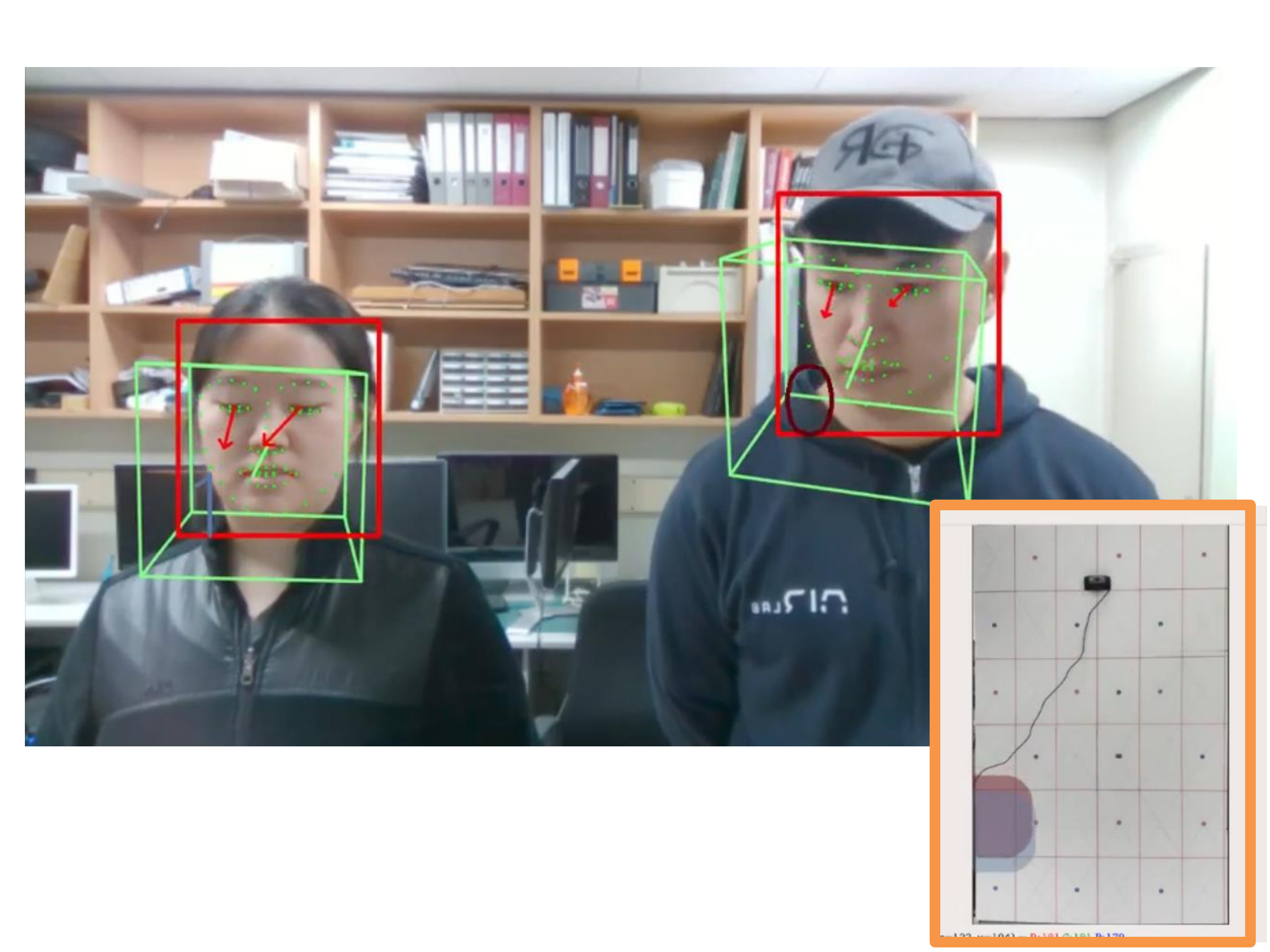}}
\caption {An example of how the Gaze Mapping System works in a test environment. Figure shows the area that two users stare at. each parts were displayed in a different color on the panel.(USER CASE 1)}
\label{fig:11}
\end{figure}

\section{Conclusions}
Gaze Mapping is a study that measures the movement of the eyeball to measure where a person is looking and how long the gaze stays at a certain point. Our eyes are one of the main organs used for decision making and learning and are the only way to accurately measure and understand visual attention.

The proposed non-contact gaze mapping system can map the user eye gaze without mounting a wearable hardware device, unlike existing studies.
Various applications are possible by real-time recognition and mapping of the natural gaze without the user's perception. In addition, the user can easily implement the Gaze mapping system with the edge equipment and the depth camera set for each shelf and has the advantage of excellent applicability. Therefore, Gaze Mapping System can be used extensively for researchers who study human behavior, and people who in charge of marketing related business of distribution companies.

\begin{figure*}
\begin{center}
\includegraphics[width=\textwidth,trim={0cm 0cm 0cm 0cm}]{{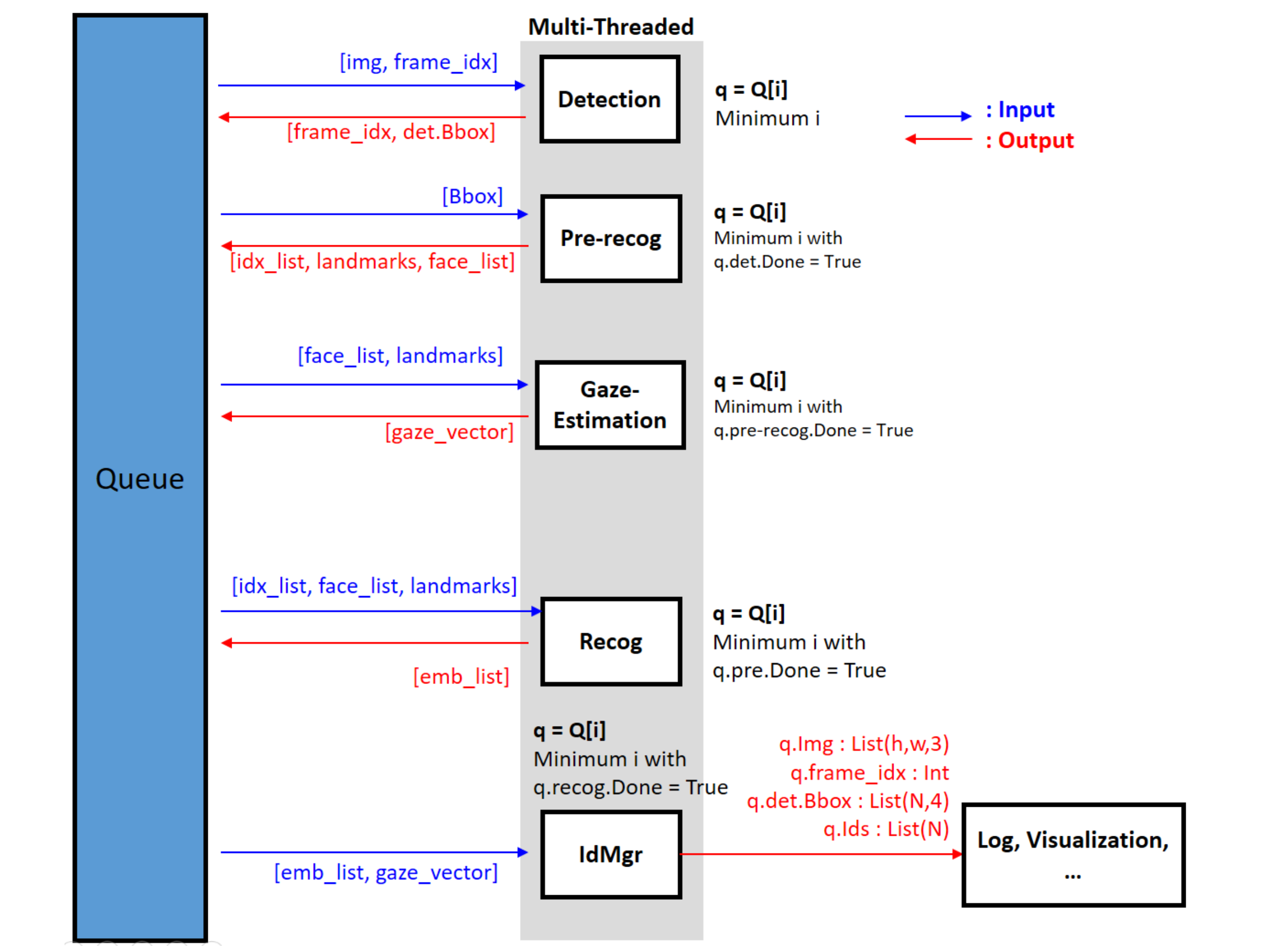}}
 \caption{Diagram of Gaze Mapping System parallelization. Based on Queue and Multi-Process, resource use is maximized to improve speed, and CPU resources can be dynamically allocated according to the load of each task to improve overall inference speed.}
\label{fig:2}
\end{center}
\end{figure*}

\section*{Acknowledgment}

The authors would like to thank...

\ifCLASSOPTIONcaptionsoff
  \newpage
\fi

\begin{IEEEbiography}[{\includegraphics[width=1in,height=1.25in,clip,keepaspectratio]{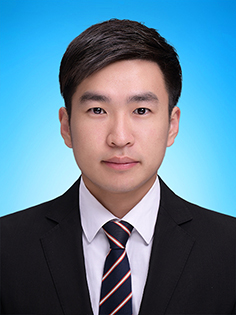}}]{Hoyeon Ahn} received the B.S. degree in industrial engineering from the Ajou University, Suwon, Republic of Korea, in 2016 and received the M.S. degrees in electrical engineering from the Kyungpook National University, Daegu, Republic of Korea, in 2018. He is a PhD candidate of the School of  Electrical Engineering and Computer Science in Gwangju Institute of Science and Technology (GIST), Gwangju, Republic of Korea. Presently, His research interests include artificial intelligence, machine Learning, pattern recognition.\end{IEEEbiography}


\end{document}